\def\year{2022}\relax
\documentclass[letterpaper]{article} 
\usepackage{aaai22}  
\usepackage{times}  
\usepackage{helvet}  
\usepackage{courier}  
\usepackage[hyphens]{url}  
\usepackage{graphicx} 
\usepackage{natbib}  
\usepackage{caption} 
\DeclareCaptionStyle{ruled}{labelfont=normalfont,labelsep=colon,strut=off} 
\usepackage{color}
\usepackage{graphicx} 
\usepackage{makecell}
\usepackage{multirow}
\usepackage{algorithm,algorithmic}
\usepackage{amsmath,amssymb,amsthm,amsfonts}
\usepackage{enumerate}
\usepackage{graphicx}
\usepackage{bm}
\usepackage{footnote}
\usepackage{url}
\usepackage{newfloat}
\usepackage{listings}
\floatstyle{ruled}
\newfloat{listing}{tb}{lst}{}
\floatname{listing}{Listing}
\title{Making Adversarial Examples More Transferable and Indistinguishable}

\author{
	Junhua Zou\textsuperscript{\rm 1},
	Yexin Duan\textsuperscript{\rm 1},
	Boyu Li\textsuperscript{\rm 2},
	Wu Zhang\textsuperscript{\rm 1},
	Yu Pan\textsuperscript{\rm 1},
	Zhisong Pan\textsuperscript{\rm 1}\thanks{Corresponding authors},
}
\affiliations{
     \textsuperscript{\rm 1}{Command and Control Engineering College, Army Engineering University, Nanjing 210007, China}\\     
     \textsuperscript{\rm 2}{School of Computer Science and Technology, Huazhong University of Science and Technology, Wuhan, 430074, China}\\
     278287847@qq.com, hotpzs@hotmail.com
}

\begin{document}

\maketitle
	
	\begin{abstract}
		Fast gradient sign attack series are popular methods that are used to generate adversarial examples. However, most of the approaches based on fast gradient sign attack series cannot balance the indistinguishability and transferability due to the limitations of the basic sign structure. To address this problem, we propose a method, called Adam Iterative Fast Gradient Tanh Method (AI-FGTM), to generate indistinguishable adversarial examples with high transferability. Besides, smaller kernels and dynamic step size are also applied to generate adversarial examples for further increasing the attack success rates. Extensive experiments on an ImageNet-compatible dataset show that our method generates more indistinguishable adversarial examples and achieves higher attack success rates without extra running time and resource. Our best transfer-based attack NI-TI-DI-AITM can fool six classic defense models with an average success rate of 89.3\% and three advanced defense models with an average success rate of 82.7\%, which are higher than the state-of-the-art gradient-based attacks. Additionally, our method can also reduce nearly 20\% mean perturbation. We expect that our method will serve as a new baseline for generating adversarial examples with better transferability and indistinguishability.
	\end{abstract}
	
	\section{Introduction}
	
	Despite the great success on many tasks, deep neural networks (DNNs) have been shown that they are vulnerable to adversarial examples \cite{DBLP:journals/corr/GoodfellowSS14,DBLP:journals/corr/SzegedyZSBEGF13}, i.e., the inputs with imperceptible perturbations can cause the incorrect results of DNNs. Moreover, a tougher problem terms transferability \cite{DBLP:conf/iclr/LiuCLS17,DBLP:conf/cvpr/Moosavi-Dezfooli17} that the adversarial examples crafted by a known DNN can also fool other unknown DNNs. Consequently, adversarial examples present severe threats to the real-world applications \cite{DBLP:conf/icml/AthalyeEIK18,DBLP:conf/cvpr/EykholtEF0RXPKS18,DBLP:conf/iclr/KurakinGB17a} and have motivated extensive research on defense methods \cite{DBLP:conf/iclr/MadryMSTV18,DBLP:conf/cvpr/LiaoLDPH018,DBLP:conf/iclr/GuoRCM18,DBLP:conf/iclr/RaghunathanSL18,DBLP:conf/icml/WongK18,DBLP:conf/icml/PangDZ18,DBLP:conf/iclr/SamangoueiKC18}. Foolbox \cite{DBLP:journals/corr/RauberBB17} roughly categorizes attack methods into three types: the gradient-based methods \cite{DBLP:conf/cvpr/DongLPS0HL18,DBLP:journals/corr/GoodfellowSS14,DBLP:conf/iclr/KurakinGB17}, the score-based methods \cite{DBLP:journals/corr/NarodytskaK16}, and the decision-based methods \cite{DBLP:conf/iclr/BrendelRB18,DBLP:journals/corr/abs-1904-02144}. In this paper, we focus on the gradient-based methods. Although the adversarial examples crafted by using the gradient-based methods satisfy the ${{L}_p}$ bound and continually achieve higher black-box success rates, these examples can be identified easily. In addition, the adversarial examples generated by the approaches based on the basic sign structure are limited. Taking TI-MI-FGSM (the combination of translation-invariant method \cite{DBLP:conf/cvpr/DongPSZ19} and momentum iterative fast gradient sign method \cite{DBLP:conf/cvpr/DongLPS0HL18}) as an example, the gradient processing steps, such as Gaussian blur, the gradient normalization, and the sign function, severely damage the gradient information. Additionally, the sign function also increases the perturbation size.
	
	\begin{figure*}[!t]
		\centering
		\includegraphics[width=1\textwidth]{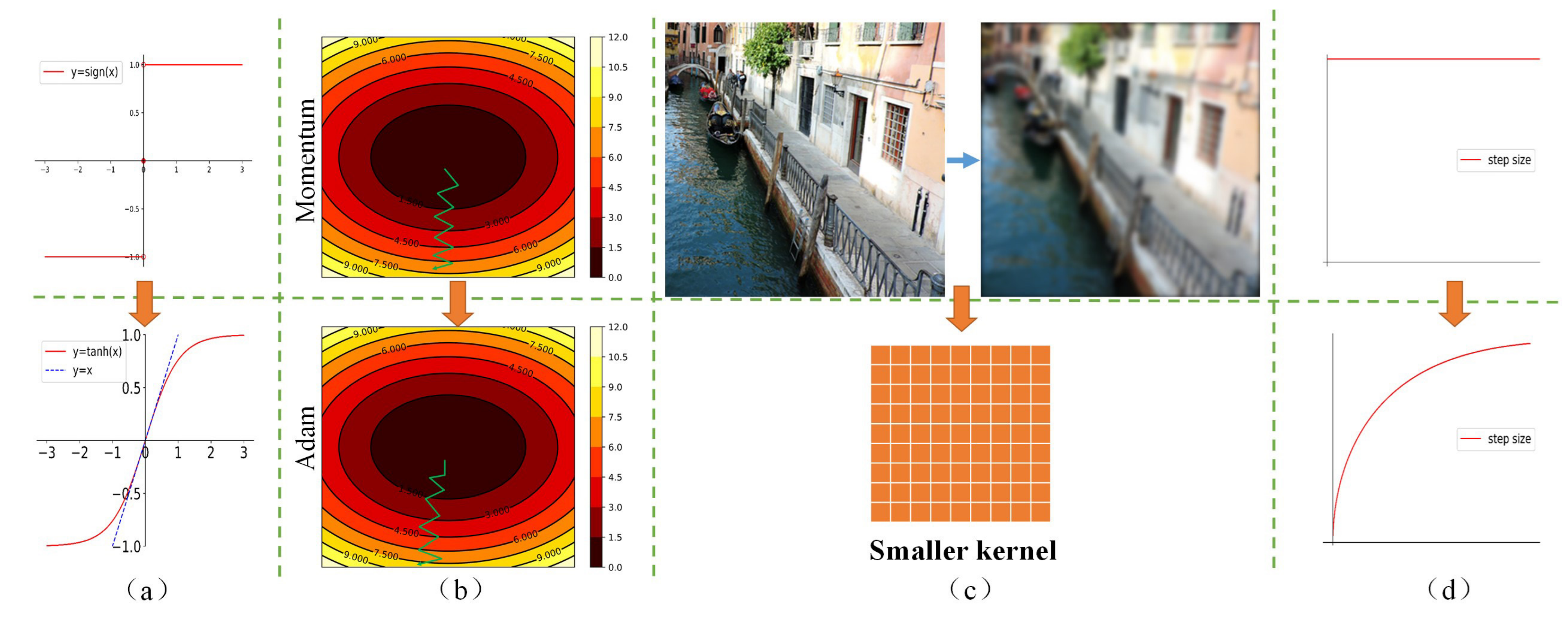}
		\caption{Overview of the our method. (a) We replace the sign function with the tanh function to generate smaller perturbations. (b) We use Adam instead of momentum method and gradient normalization to get larger losses in only ten iterations. (c) We use smaller kernels in Gaussian blur to avoid the loss of gradient information. (d) We gradually increase the step size.}
		\label{fig:1}
	\end{figure*}
	
	\begin{figure*}[!t]
		\centering
		\includegraphics[width=1\textwidth]{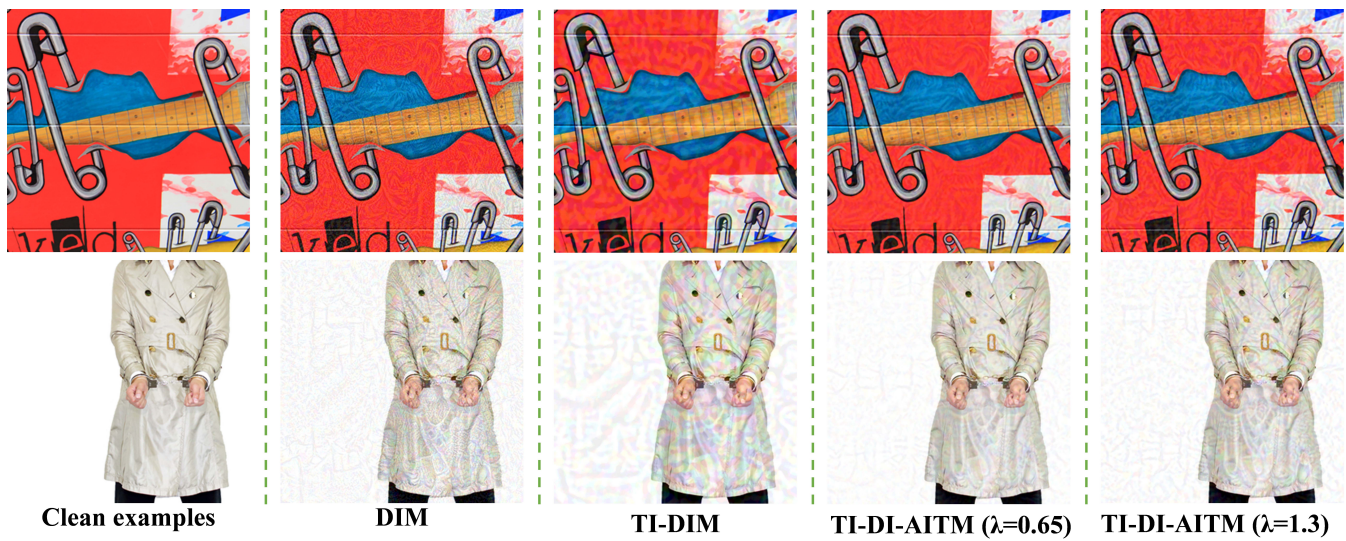}
		\caption{The comparison of clean examples and adversarial examples crafted by three combined attacks. (1) TI-DI-AITM ($\lambda$=0.65) achieves similar success rates as TI-DIM and generates much more indistinguishable adversarial examples. (2) TI-DI-AITM ($\lambda$=1.3) achieves much higher success rates than TI-DIM and generates more indistinguishable adversarial examples.}
		\label{fig:2}
	\end{figure*}
	
	In this paper, we propose a method, called \textbf{Adam Iterative Fast Gradient Tanh Method (AI-FGTM)}, which improves the indistinguishability and transferability of adversarial examples. It is known that the fast gradient sign attack series iteratively process gradient information with transformation, normalization, and the sign function. To preserve the gradient information as much as possible, AI-FGTM modifies the major gradient processing steps. Still take TI-MI-FGSM as an example, to avoid the loss of gradient information and generate imperceptible perturbations, we replace the momentum algorithm and the sign function with Adam \cite{DBLP:journals/corr/KingmaB14} and the tanh function, respectively. Then, we employ dynamic step size and smaller filters in Gaussian blur. The overview of AI-FGTM is shown in Fig.~\ref{fig:1}, and the detailed process will be given in \textbf{Methodology}. Furthermore, combining the existing attack methods with AI-FGTM can get much smaller perturbations and deliver state-of-the-art success rates. Fig.~\ref{fig:2} shows the comparison of different examples, where the adversarial examples are crafted by three combined attacks, namely, DIM \cite{DBLP:conf/cvpr/XieZZBWRY19}, TI-DIM \cite{DBLP:conf/cvpr/DongPSZ19} and TI-DI-AITM (the combination of TI-DIM and our method).


	In summary, we make the following contributions:
	
	\begin{itemize}
		\item[1.] Inspired by the limitations of the fast gradient sign series, we propose AI-FGTM in which the major gradient processing steps are improved to boost the indistinguishability and transferability of adversarial examples.
		\item[2.] We show that AI-FGTM integrated with other transfer-based attacks can obtain much smaller perturbations and larger losses than the current sign attack series.
		\item[3.] The empirical experiments prove that without extra running time and resource, our best attack fools six classic defense models with an average success rate of 89.3\% and three advanced defense models with an average success rate of 82.7\%, which are higher than the state-of-the-art gradient-based attacks.
	\end{itemize}
	
	\section{Review of Existing Attack Methods}
	
	\subsection{Problem definition}
	Let $\left\{ {{{\left( {{f_{Wi}}} \right)}_{i \in \left[ N \right]}},{{\left( {{f_{Bj}}} \right)}_{j \in \left[ M \right]}}} \right\}$ be a set of pre-trained classifiers, where $\ {{\left( {{f_{Wi}}} \right)}_{i \in \left[ N \right]}}{\rm{ }}$ denotes the white-box classifiers and ${{\left( {{f_{Bj}}} \right)}_{j \in \left[ M \right]}}$ represents the unknown classifiers. Given a clean example $x$, it can be correctly classified to the ground-truth label ${y^{true}}$ by all pre-trained classifiers. It is possible to craft an adversarial example ${x^{adv}}$ that satisfies ${\left\| {{x^{adv}} - x} \right\|_p} \le \varepsilon $ by using the white-box classifiers, where $p$ could be $0$, $1$, $2$, $\infty $, and $\varepsilon $ is the perturbation size. In this paper, we focus on non-targeted attack with $p = \infty$. Note that, the adversarial example ${x^{adv}}$ can mislead the white-box classifiers and the unknown classifiers simultaneously.
	
	\subsection{The gradient-based methods}
	Here, we introduce the family of the gradient-based methods.
	
	Fast Gradient Sign Method (FGSM) \cite{DBLP:journals/corr/GoodfellowSS14} establishes the basic framework of the gradient-based methods. It efficiently crafts an adversarial example ${x^{adv}}$ by using one-step update while maximizing the loss function $J( {{x^{adv}},{y^{true}}})$ of a given classifier as
	\begin{equation}\label{eq:4}
	{x^{adv}} = x + \varepsilon  \cdot {\rm{sign}}( {{\nabla _x}J( {x,{y^{true}}} )} ),
	\end{equation}
	where ${\nabla _x}J( {\cdot,\cdot})$ computes the gradient of the loss function w.r.t. $x$, ${\rm{sign}}( \cdot )$ is the sign function, and $\varepsilon$ is the given scalar value that restricts the ${L_\infty}$ norm of the perturbation.
	
	Basic Iterative Method (BIM) \cite{DBLP:conf/iclr/KurakinGB17a} is the iterative version of FGSM that performs better in white-box attack but less effective in transfer-based attack. It iteratively updates the adversarial example $x_t^{adv}$ with a small step size ${\rm{\alpha }}$ as
	\begin{equation}\label{eq:6}
	x_{t + 1}^{adv} = {\rm{Clip}}_\varepsilon ^x\left\{ {x_t^{adv} + \alpha \cdot{\rm{sign}}( {{\nabla _{x_t^{adv}}}J( {x_t^{adv},{y^{true}}})})} \right\},
	\end{equation}
	where ${\rm{\alpha }} = \varepsilon /T$ with $T$ denoting the number of iterations. ${\rm{Clip}}_\varepsilon ^x\left\{  \cdot  \right\}$ performs per-pixel clipping as
	\begin{equation}\label{eq:5}
	{\rm{Clip}}_\varepsilon ^x\left\{ {x'} \right\} = \min \left\{ {255,x + \varepsilon ,\max \left\{ {0,x - \varepsilon ,x'} \right\}} \right\}.
	\end{equation}
	
	Momentum Iterative Fast Gradient Sign Method (MI-FGSM) \cite{DBLP:conf/cvpr/DongLPS0HL18} enhances the transferability of adversarial examples by incorporating momentum term into gradient process, given as
	\begin{equation}\label{eq:9}
	{g_{t + 1}} = \mu  \cdot {g_t} + \frac{{{\nabla _{x_t^{adv}}}J( {x_t^{^{adv}},{y^{true}}} )}}{{{{\left\| {{\nabla _{x_t^{adv}}}J( {x_t^{^{adv}},{y^{true}}})} \right\|}_1}}},
	\end{equation}
	\begin{equation}\label{eq:10}
	x_{t + 1}^{^{adv}} = {\rm{Clip}}_\varepsilon ^x\left\{{x_t^{^{adv}} + \alpha  \cdot {\rm{sign}}( {{g_{t + 1}}} )} \right\},
	\end{equation}
	where ${g_{t + 1}}$ denotes the accumulated gradient at ${(t+1)}_{th}$ iteration , and $\mu$ is the decay factor of ${g_{t + 1}}$.
	
	Nesterov Iterative Method (NIM) \cite{DBLP:journals/corr/abs-1908-06281} integrates an anticipatory update into MI-FGSM and further increases the transferability of adversarial examples. The update procedures are expressed as
	\begin{equation}\label{eq:11}
	x_t^{nes} = x_t^{adv} + \alpha \cdot \mu \cdot {g_t},
	\end{equation}
	\begin{equation}\label{eq:12}
	{g_{t + 1}} = \mu \cdot {g_t} + \frac{{{\nabla _{x_t^{nes}}}J( {x_t^{nes},{y^{true}}})}}{{{{\left\| {{\nabla _{x_t^{nes}}}J( {x_t^{nes},{y^{true}}})} \right\|}_1}}},
	\end{equation}
	\begin{equation}\label{eq:13}
	x_{t + 1}^{adv} = {\rm{Clip}}_\varepsilon ^x\left\{{x_t^{adv} + \alpha  \cdot {\rm{sign}}( {{g_{t + 1}}})} \right\}.
	\end{equation}
	
	Scale-Invariant Method (SIM) \cite{DBLP:journals/corr/abs-1908-06281} applies the scale copies of the input image to further improve the transferability. However, SIM requires much more running time and resource.
	
	Diverse Input Method (DIM) \cite{DBLP:conf/cvpr/XieZZBWRY19} applies random resizing and padding to the adversarial examples with the probability $p$ at each iteration. DIM can be easily integrated into other gradient-based methods to further boost the transferability of adversarial examples. The transformation function ${T(x_t^{adv},p)}$ is
	
	\begin{equation}\label{eq:44}
	T(x_t^{adv},p)=\left\{
	\begin{array}{ll}
	T(x_t^{adv}),p\\
	x_t^{adv},(1-p)
	\end{array} \right..
	\end{equation}
	
	Translation-Invariant Method (TIM) \cite{DBLP:conf/cvpr/DongPSZ19} optimizes an adversarial example by an ensemble of translated examples as
	\begin{equation}\label{eq:45}
	\left.\begin{array}{ll}
	x_{t+1}^{adv} = \sum_{i,j}T_{ij}(x_t^{adv}),
	s.t.\,\,\|x_t^{adv} - x^{real}\|_{\infty} \leq \epsilon,
	\end{array}\right.
	\end{equation}
	where $T_{ij}(x_t^{adv})$ denotes the translation function that respectively shifts input $x_t^{adv}$ by $i$ and $j$ pixels along the two-dimensions. And TIM calculates the gradient of the loss function at a point $\hat x_t^{adv}$, convolves the gradient with a pre-defined Gaussian filter and updates as
	\begin{equation}\label{eq:add1}
	\begin{array}{l}
	{g^{'}} = {\nabla _{x_t^{adv}}} ( {\sum\limits_{i,j} {{w_{ij}}J ( {{T_{ij}} ( {x_t^{adv}} ),{y^{true}}} )} } ){|_{x_t^{adv} = \hat x_t^{adv}}}\\
	\quad \approx \sum\limits_{i,j} {{w_{ij}}{T_{ - i - j}} ( {{\nabla _{x_t^{adv}}}J ( {x_t^{adv},{y^{true}}} )}  ){|_{x_t^{adv} = \hat x_t^{adv}}}} \\
	\quad\approx W*{\nabla _{x_t^{adv}}}J ( {x_t^{adv},{y^{true}}} )
	\end{array},
	\end{equation}

	\begin{equation}\label{eq:46}
	x_{t + 1}^{adv} = {\rm{Clip}}_\varepsilon ^x\left\{{x_t^{adv} + \alpha  \cdot {\rm{sign}}( {{g^{'}}})} \right\}.
	\end{equation}
	
	Note that, with limited running time and computing resources, the combination of NIM, TIM and DIM (NI-TI-DIM) is strong transfer-based attack method so far.
	
	
	
	
	\section{Methodology}
	\label{s3}
	\subsection{Motivations}
	
	Based on the contradiction that the adversarial examples achieve high success rates but can be identified easily, our observations are shown as follows:
	
	\begin{itemize}
		\item[1.] The sign function in gradient-based methods has two main disadvantages. On the one hand, the sign function normalizes all the gradient values to 1, -1 or 0, and thus leads to the loss of gradient information. On the other hand, the sign function normalizes some small gradient values to 1 or -1, which increases the perturbation size. While the tanh function not only normalizes the large gradient values as the sign function, but also maintains the small gradient values as function $y=x$. Therefore, the tanh function can replace the sign function and reduce the perturbation size.
		\item[2.] With iterations $T=10$, the applications of Nesterov’s accelerated gradient (NAG) \cite{DBLP:journals/corr/abs-1908-06281} and the momentum algorithm \cite{DBLP:conf/cvpr/DongLPS0HL18} in adversarial attacks demonstrate that we can migrate other methods to generate adversarial examples. Moreover, the $t_{\rm{th}}$ gradient ${\nabla _{x_t^{adv}} J( {x_t^{adv},{y^{true}}})}$ is normalized by the $L_1$ distance of itself before the momentum algorithm. Intuitively, due to the performance of traditional convergence algorithms, Adam can achieve larger losses than the momentum algorithm in such small number of iterations. Additionally, Adam can normalize the gradient with ${{{m_t}} \mathord{\left/
				{\vphantom {{{m_t}} {\sqrt {{v_t} + \delta } }}} \right.
				\kern-\nulldelimiterspace} {\sqrt {{v_t} + \delta } }}$, where ${m_t}$ denotes the first moment vector, ${v_t}$ is the second moment vector and $\delta  = {10^{ - 8}}$.
		\item[3.] Traditional convergence algorithms employ learning rate decay to improve the model performance. Existing gradient-based methods set stable step size ${\rm{\alpha }} = \varepsilon /T$. In intuition, we can improve the transferability with the step size change. Different from the traditional convergence algorithms, the attack methods with the $\varepsilon $-ball restriction aim to maximize the loss function of the target models. Hence, we use the increasing step size with $\sum\nolimits_{t = 0}^{{\rm{T - 1}}} {{\alpha _{\rm{t}}}}  = {\rm{\varepsilon }}$.
		\item[4.] Dong et al. \cite{DBLP:conf/cvpr/DongPSZ19} show that Gaussian blur with a large kernel improves the transferability of adversarial examples. However, Gaussian blur with larger kernels leads to the loss of the gradient information. Using the modifications mentioned above, the gradient information is preserved and plays a more important role in generating adversarial examples. Consequently, we apply smaller kernels in Gaussian blur to avoid the loss of the gradient information.
	\end{itemize}
	
	
	Based on the above four observations, we propose AI-FGTM to craft the adversarial examples which are expected to be more transferable and indistinguishable.
	
	\subsection{AI-FGTM}
	Adam \cite{DBLP:journals/corr/KingmaB14} uses the exponential moving averages of squared past gradients to mitigate the rapid decay of learning rate. Essentially, this algorithm limits the reliance of update to only the past few gradients by the following simple recursion:
	
	\begin{equation}\label{eq:14}
	{m_{t + 1}} = {\beta _1}{m_t} + ({1 - {\beta _1}}){g_{t + 1}},
	\end{equation}
	
	\begin{equation}\label{eq:114}
	{v_{t + 1}} = {\beta _2}{v_t} +( {1 - {\beta _2}})g_{t + 1}^2,
	\end{equation}
	
	\begin{equation}\label{eq:15}
	{\theta _{t + 1}} = {\theta _t} - \alpha  \cdot \frac{{\sqrt {( {1 - \beta _2^t})} }}{{1 - \beta _1^t}} \cdot \frac{{{m_{t + 1}}}}{{\sqrt {{v_{t + 1}} + \delta } }},
	\end{equation}
	where ${m_t}$ denotes the first moment vector, ${v_t}$ represents the second moment vector, ${\beta_1}$ and ${\beta_2}$ are the exponential decay rates.
	
	\begin{table*}[t]
		\caption{Abbreviations used in the paper}
		\label{table1}
		\centering
		\begin{tabular}{|l|l|l|}
			\hline
			Abbreviation     & Definition \\
			\hline
			\hline
			TI-DIM     & The combination of MI-FGSM, TIM and DIM \\
			NI-TI-DIM  & The combination of MI-FGSM, NIM, TIM and DIM \\
			SI-NI-TI-DIM & The combination of MI-FGSM, NIM, SIM, TIM and DIM\\
			TI-DI-AITM & The combination of AI-FGTM, TIM and DIM\\
			NI-TI-DI-AITM & The combination of AI-FGTM, NIM, TIM and DIM\\
			SI-NI-TI-DI-AITM & The combination of AI-FGTM, NIM, SIM, TIM and DIM\\
			\hline
		\end{tabular}
	\end{table*}

	\begin{table*}[!t]
	\caption{The running time (s) of generating 1000 adversarial examples for Inc-v3, Inc-v4, IncRes-v2, Res-v2-101 and the ensemble of theses four models.}
	\label{table5}
	\centering
	\begin{tabular}{|c|ccccc|}
		\hline
		Attack     & Inc-v3 & Inc-v4 & IncRes-v2 & Res-v2-101 & Model ensemble \\
		\hline
		\hline
		TI-DIM     & 172.8 & 261.2 & 277.8 & 234.0 & 767.5 \\
		NI-TI-DIM  & 174.5 & 238.9 & 291.8 & 243.0 & 830.2 \\
		SI-NI-TI-DIM  & 608.2 & 1086.3 & 1156.2 & 1096.2 & 3490.2 \\
		TI-DI-AITM & 170.6 & 258.5 & 280.4 & 239.3 & 762.7 \\
		NI-TI-DI-AITM & 173.5 & 253.7 & 288.1 & 242.1 & 770.1 \\
		SI-NI-TI-DI-AITM  & 603.6 & 1103.9 & 1119.4 & 1123.1 & 3341.6 \\
		\hline
	\end{tabular}
	\end{table*}
	
	Due to the opposite optimization objectives, we apply Adam into adversarial attack with some modifications. Starting with $x_0^{adv} = x$, $m_0 = 0$ and $v_0 = 0$, the first moment estimate and the second moment estimate are presented as follows:
	
	\begin{equation}\label{eq:16}
	{m_{t + 1}} = {m_t} + {\mu _1} \cdot {\nabla _{x_t^{adv}}}J( {x_t^{adv},{y^{true}}}),
	\end{equation}
	
	\begin{equation}\label{eq:116}
	{v_{t + 1}} = {v_t} + {\mu _2} \cdot {( {{\nabla _{x_t^{adv}}}J( {x_t^{adv},{y^{true}}})})^2},
	\end{equation}
	where ${\mu_1}$ and ${\mu_2}$ denote the first moment factor and second moment factor, respectively. We replace the sign function with the tanh function and update $x_{t + 1}^{adv}$ as
	\begin{equation}\label{eq:17}
	{\alpha}_t  = \frac{\varepsilon }{{\mathop \sum \nolimits_{t = 0}^{T - 1} \frac{{1 - \beta _1^{t + 1}}}{{\sqrt {( {1 - \beta _2^{t + 1}} )} }}}}\frac{{1 - \beta _1^{t + 1}}}{{\sqrt {( {1 - \beta _2^{t + 1}} )} }},
	\end{equation}
	
	\begin{equation}\label{eq:18}
	x_{t + 1}^{adv} = {\rm{Clip}}_\varepsilon ^x\left\{ {x_t^{adv} + {\alpha _t}\cdot{\rm{tanh}}( {\lambda \frac{{{m_{t + 1}}}}{{\sqrt {{v_{t + 1}}}  + \delta }}} )} \right\},
	\end{equation}
	where ${\beta_1}$ and ${\beta_2}$ are exponential decay rates, and $\lambda $ denotes the scale factor. Specifically, ${\alpha}_t $ is the increasing step size with $\sum\nolimits_{t = 0}^{{\rm{T - 1}}} {{\alpha _{\rm{t}}}}  = {\rm{\varepsilon }}$. Then the tanh function reduces the perturbations of adversarial examples without any success rate reduction. Furthermore, ${m_{t + 1}}/( {\sqrt {{v_{t + 1}}}  + \delta } )$ replaces the $L_1$ normalization and the first moment estimate of Eq.~\ref{eq:9} due to the fact that Adam has faster divergence speed than momentum attack algorithm (as shown in Fig.~\ref{fig:1}(b)).
	
	\subsection{ The combination of AI-FGTM and NIM}
	
	Inspired by NIM that integrates an anticipatory update into MI-FGSM. Similarly, we can also integrate an anticipatory update into AI-FGTM. We first calculate the step size in each iteration as Eq.~\ref{eq:17}, and the Nesterov term can be expressed as
	
	\begin{equation}\label{eq:21}
	x_t^{nes} = x_t^{adv} + {\alpha}_t \cdot {\frac{{{m_{t}}}}{{\sqrt {{v_{t}}}  + \delta }}}.
	\end{equation}
	
	The remaining update procedures are similar to Eq.~\ref{eq:16}, Eq.~\ref{eq:116} and Eq.~\ref{eq:18}, which can be expressed as
	
	\begin{equation}\label{eq:22}
	{m_{t + 1}} = {m_t} + {\mu _1} \cdot {\nabla _{x_t^{nes}}}J( {x_t^{nes},{y^{true}}} ),
	\end{equation}
	
	\begin{equation}\label{eq:23}
	{v_{t + 1}} = {v_t} + {\mu _2} \cdot {( {{\nabla _{x_t^{nes}}}J( {x_t^{nes},{y^{true}}})})^2},
	\end{equation}
	
	\begin{equation}\label{eq:24}
	x_{t + 1}^{adv} = {\rm{Clip}}_\varepsilon ^x\left\{ {x_t^{adv} + {\alpha _t}\cdot{\rm{tanh}}( {\lambda \frac{{{m_{t + 1}}}}{{\sqrt {{v_{t + 1}}}  + \delta }}} )} \right\}.
	\end{equation}
	
	We summarize NI-TI-DI-AITM as the combination of AI-FGTM, NIM, TIM and DIM, and the procedure is given in Algorithm~\ref{al:2}.
	
	\begin{algorithm}[ht]
		\caption{NI-TI-DI-AITM}
		\label{al:2}
		{\bf Input:}
		A clean example $x$ and its ground-truth label ${y^{true}}$;\\
		{\bf Parameters:}
		The perturbation size $\varepsilon$; the iteration number $T$; the decay factors ${\mu _1}$ and ${\mu _2}$; the exponential decay rates ${\beta_1}$ and ${\beta_2}$; the scale factor $\lambda$; the probability $p$. \\
		{\bf Output:}
		An adversarial example $x^{adv}$.\\
		\begin{algorithmic}[1]
			\STATE ${m_0} = 0$; ${v_0} = 0$; $x_{_0}^{adv} = x$;
			\FOR {$t = 0$ to $T-1$}
			\STATE Input $x_t^{adv}$;
			\STATE Update step size $\alpha$ by Eq.~\ref{eq:17};
			\STATE Obtain the Nesterov term $x_t^{nes}$ by Eq.~\ref{eq:21};
			\STATE Obtain the diverse input $T(x_t^{nes},p)$ by Eq.~\ref{eq:44};
			\STATE Compute the gradient ${\nabla _{x_t^{adv}}}J({T(x_t^{nes},p),{y^{true}}})$;
			\STATE Obtain processed gradient $g'$ by Eq.~\ref{eq:add1};
			\STATE Update ${m_{t + 1}}$ by ${m_{t + 1}} = {m_t} + {\mu _1} \cdot g'$;
			\STATE Update ${v_{t + 1}}$ by ${v_{t + 1}} = {v_t} + {\mu _2}\cdot{( {g'} )^2}$;
			\STATE Update $x_{t + 1}^{adv}$ by Eq.~\ref{eq:24};
			\ENDFOR
			\STATE {\bf Return} ${x^{adv}} = x_t^{adv}$.
		\end{algorithmic}
	\end{algorithm}
	
	\section{Experiments}
	In this section, we provide extensive experimental results on an ImageNet-compatible dataset to validate our method. First, we introduce the experimental setting. Then, we compare the running efficiency of different transfer-based attacks. Next, we present the ablation study of the effects of different parts of our method. Finally, we compare the results of the baseline attacks. Table ~\ref{table1} presents the definitions of the abbreviations used in this paper.
	
	\subsection{Experimental setting}
	\label{sec4}
	
	\textbf{Dataset}. We utilize 1000 images \footnote{\url{https://github.com/tensorflow/cleverhans/tree/master/examples/nips17_adversarial_competition/dataset}} which are used in the NIPS 2017 adversarial competition to conduct the following experiments.
	
	\textbf{Models}. In this paper, we employ thirteen models to perform the following experiments. Four non-defense models (Inception v3 (Inc-v3) \cite{DBLP:conf/cvpr/SzegedyVISW16}, Inception v4 (Inc-v4), Inception ResNet v2 (IncRes-v2) \cite{DBLP:conf/aaai/SzegedyIVA17}, and ResNet v2-101 (Res-v2-101) \cite{DBLP:conf/eccv/HeZRS16}) are used as white-box models to craft adversarial examples. Six defense models (Inc-v3ens3, Inc-v3ens4, IncResv2ens \cite{DBLP:conf/iclr/TramerKPGBM18}, high-level representation guided denoiser (HGD) \cite{DBLP:conf/cvpr/LiaoLDPH018}, input transformation through random resizing and padding (R\&P) \cite{DBLP:conf/iclr/XieWZRY18}, and rank-3 submission \footnote{\url{https://github.com/anlthms/nips-2017/tree/master/mmd}} in the NIPS 2017 adversarial competition) are employed as classic models to evaluate the crafted adversarial examples. In addition, we also evaluate the attacks with three advanced defenses (Feature Distillation \cite{DBLP:conf/cvpr/LiuLLXLWW19}, Comdefend \cite{DBLP:conf/cvpr/JiaWCF19}, and Randomized Smoothing \cite{DBLP:conf/icml/CohenRK19}).
	
	\textbf{Baselines}. We focus on the comparison of TI-DIM, NI-TI-DIM, TI-DI-AITM and NI-TI-DI-AITM, where TI-DIM and NI-TI-DIM are both the state-of-the-art methods.
	
	\textbf{Hyper-parameters}. According to TI-DIM \cite{DBLP:conf/cvpr/DongPSZ19} and NI-FGSM \cite{DBLP:journals/corr/abs-1908-06281}, we set the maximum perturbation ${\rm{\varepsilon }} = 16$, and the number of iteration $T = 10$. Specifically, we set the kernel size to $15 \times 15$ in normal TI-DIM and NI-TI-DIM while $9 \times 9$ in TI-DI-AITM. The exploration of appropriate settings of our method is illustrated in \textbf{Appendix}.

	\textbf{The mean perturbation size}. For an adversarial example $x^{adv}$ with the size of ${M \times N \times 3}$, the mean perturbation size ${P_m}$ can be calculated as 
	\begin{equation}\label{eq:mean}
	{P_m} = \frac{{\mathop \sum \nolimits_{i = 1}^M \mathop \sum \nolimits_{j = 1}^N \mathop \sum \nolimits_{k = 1}^3 \left| {x_{ijk}^{adv} - {x_{ijk}}} \right|}}{{M \times N \times 3}},
	\end{equation}
	where $x_{ijk}$ denotes the value of channel $k$ of the image $x$ at coordinates $( {i,j})$.
	
	\begin{table*}[!t]
		\caption{Ablation study of the effects of the tanh function, the Adam optimizer, the kernel size and dynamic step size. The adversarial examples are generated for Inc-v3, Inc-v4, IncRes-v2, Res-v2-101 respectively using TI-DIM and TI-DIM with different parts of our method. We compare the mean perturbations and the mean attack success rates of the generated adversarial examples against six classic defense models.}
		\label{table_ablation}
		\centering
		\begin{tabular}{|c|cccc|cc|}
			\hline
			Attack                   & tanh & Adam & smaller kernels & dynamic step size & mean success rate (\%) & mean perturbation \\ \hline
			\hline
			\multirow{6}{*}{TI-DIM} &      &      &                &                   & 82.0                & 10.46             \\ \cline{2-7} 
			& \checkmark    &      &                &                   & 82.4              & 9.14              \\ \cline{2-7} 
			&      & \checkmark    &                &                   & 83.6              & 9.20               \\ \cline{2-7} 
			& \checkmark    & \checkmark    &                &                   & 83.1              & 7.86              \\ \cline{2-7} 
			& \checkmark    & \checkmark    & \checkmark              &                   & 86.5              & 7.82              \\ \cline{2-7} 
			& \checkmark    & \checkmark    & \checkmark              & \checkmark                 & 88.0                & 8.11              \\ \hline	\end{tabular}
	\end{table*}
	
	\begin{table*}[t]
		\centering
		\caption{\label{table3}The success rates (\%) of adversarial attacks against six defense models under single-model setting. The adversarial examples are generated for Inc-v3, Inc-v4, IncRes-v2, Res-v2-101 respectively using TI-DIM, NI-TI-DIM, TI-DI-AITM and NI-TI-DI-AITM.}
		\begin{tabular}{|c|c|cccccc|}
			\hline
			& Attack & Inc-v3ens3 & Inc-v3ens4 & IncRes-v2ens & HGD & R\&P & NIPS-r3\\
			\hline
			\hline
			Inc-v3 & \begin{tabular}[c]{@{}c@{}}TI-DIM\\TI-DI-AITM\\ NI-TI-DIM\\ NI-TI-DI-AITM\end{tabular} & \begin{tabular}[c]{@{}c@{}}46.5\\ \textbf{53.8}\\ 48.3\\\textbf{51.9}\end{tabular} & \begin{tabular}[c]{@{}c@{}}47.3\\ \textbf{53.3}\\ 48.6\\ \textbf{52.4}\end{tabular} & \begin{tabular}[c]{@{}c@{}}38.1\\ \textbf{39.0}\\ 36.9\\ \textbf{38.2}\end{tabular}
			& \begin{tabular}[c]{@{}c@{}} 38.0\\ \textbf{40.2}\\ 37.3\\\textbf{39.3}\end{tabular} & \begin{tabular}[c]{@{}c@{}} 36.9\\ \textbf{39.1}\\ 36.8\\ \textbf{38.1}\end{tabular} & \begin{tabular}[c]{@{}c@{}} 41.1\\ \textbf{45.7}\\ 42.5\\\textbf{44.6}\end{tabular}\\
			\hline
			Inc-v4 & \begin{tabular}[c]{@{}c@{}}TI-DIM\\TI-DI-AITM\\ NI-TI-DIM\\ NI-TI-DI-AITM\end{tabular} & \begin{tabular}[c]{@{}c@{}}48.2\\ \textbf{53.2}\\ 52.4\\ \textbf{54.8}\end{tabular} & \begin{tabular}[c]{@{}c@{}}47.9\\ \textbf{51.8}\\ 51.8\\ \textbf{53.7}\end{tabular} & \begin{tabular}[c]{@{}c@{}}39.1\\ \textbf{42.4}\\ 41.3\\  \textbf{41.7}\end{tabular}
			& \begin{tabular}[c]{@{}c@{}} 40.6\\ \textbf{43.7}\\ 41.9\\  \textbf{43.9}\end{tabular} & \begin{tabular}[c]{@{}c@{}} 39.3\\ \textbf{42.5}\\ 41.1\\  \textbf{43.2}\end{tabular} & \begin{tabular}[c]{@{}c@{}} 41.5\\ \textbf{44.6}\\ 42.7\\  \textbf{44.1}\end{tabular}\\
			\hline
			IncRes-v2 & \begin{tabular}[c]{@{}c@{}}TI-DIM\\TI-DI-AITM\\ NI-TI-DIM\\ NI-TI-DI-AITM\end{tabular} & \begin{tabular}[c]{@{}c@{}}60.8\\ \textbf{64.9}\\ 61.5\\  \textbf{66.5}\end{tabular} & \begin{tabular}[c]{@{}c@{}}59.6\\ \textbf{61.8}\\ 60.4\\ \textbf{63.8}\end{tabular} & \begin{tabular}[c]{@{}c@{}}59.3\\ \textbf{62.1}\\ 59.9\\  \textbf{62.0}\end{tabular}
			& \begin{tabular}[c]{@{}c@{}}58.4\\ \textbf{62.7}\\ 60.1\\  \textbf{63.2}\end{tabular} & \begin{tabular}[c]{@{}c@{}} 60.7\\ \textbf{64.8}\\ 62.2\\\textbf{65.6}\end{tabular} & \begin{tabular}[c]{@{}c@{}} 61.3\\ \textbf{65.1}\\ 63.1\\  \textbf{65.8}\end{tabular}\\
			\hline
			Res-v2-101 & \begin{tabular}[c]{@{}c@{}}TI-DIM\\TI-DI-AITM\\ NI-TI-DIM\\ NI-TI-DI-AITM\end{tabular} & \begin{tabular}[c]{@{}c@{}}56.1\\ \textbf{62.8}\\ 59.5\\  \textbf{64.0}\end{tabular} & \begin{tabular}[c]{@{}c@{}}55.4\\ \textbf{62.8}\\ 57.7\\  \textbf{61.0}\end{tabular} & \begin{tabular}[c]{@{}c@{}}49.8\\ \textbf{54.4}\\ 50.4\\  \textbf{54.6}\end{tabular}
			& \begin{tabular}[c]{@{}c@{}} 51.3\\ \textbf{55.3}\\ 51.9\\  \textbf{54.8}\end{tabular} & \begin{tabular}[c]{@{}c@{}} 50.4\\ \textbf{54.2}\\ 50.8\\  \textbf{53.4}\end{tabular} & \begin{tabular}[c]{@{}c@{}} 52.3\\ \textbf{57.1}\\ 54.6\\  \textbf{57.6}\end{tabular}\\
			\hline
		\end{tabular}
	\end{table*}

	\begin{figure*}[t]
	\centering
	\includegraphics[width=1\textwidth]{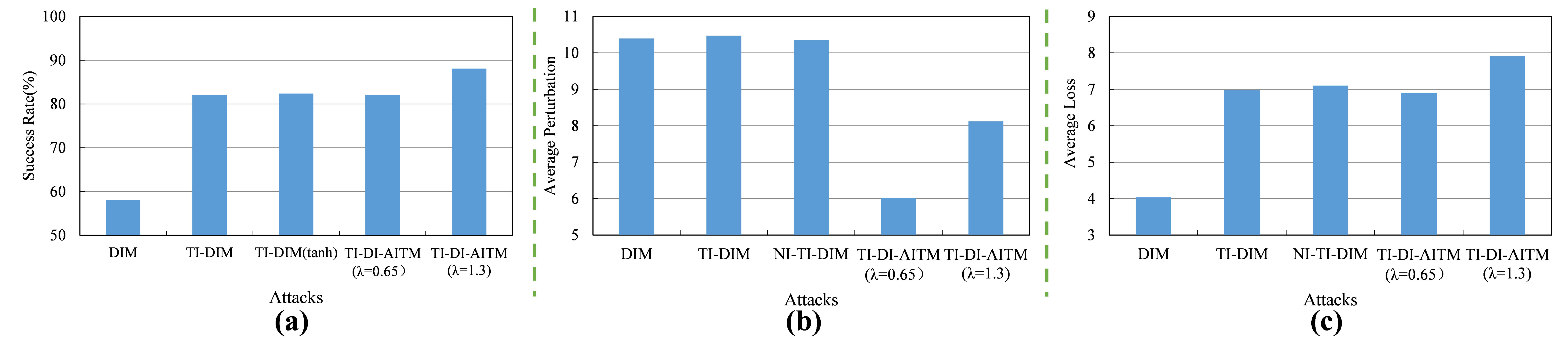}
	\caption{Results of adversarial examples generated for the ensemble of Inc-v3, Inc-v4, IncRes-v2, and Res-v2-101 using different attacks.}
	\label{fig:4}
    \end{figure*}	
	\begin{table*}[t]
		\centering
		\caption{The success rates (\%) of adversarial attacks against six defense models under model ensemble setting. The adversarial examples are generated for the ensemble of Inc-v3, Inc-v4, IncRes-v2, and Res-v2-101 using TI-DIM, NI-TI-DIM, TI-DI-AITM and NI-TI-DI-AITM.}
		\label{table4}
		\begin{tabular}{|c|cccccc|c|}
			\hline
			Attack     & Inc-v3ens3 & Inc-v3ens4 & IncRes-v2ens & HGD  & R\&P & NIPS-r3 & Average\\
			\hline
			\hline
			TI-DIM     & 83.9       & 83.2       & 78.4         & 81.9 & 81.2 & 83.6   &  82.0\\
			TI-DI-AITM & \textbf{90.2}       & \textbf{88.5}       & \textbf{85.4}         & \textbf{88.3} & \textbf{87.1} & \textbf{88.7} & \textbf{88.0} \\
			NI-TI-DIM  & 85.5       & 85.9       & 80.1         & 83.6 & 82.9 & 84.3   &  83.7\\
			NI-TI-DI-AITM & \textbf{91.8}       & \textbf{90.3}       & \textbf{85.8}         & \textbf{89.4} & \textbf{88.6} & \textbf{90.1} & \textbf{89.3} \\
			\hline
		\end{tabular}
	\end{table*}
	
	\begin{table*}[!htb]
		\centering
		\caption{The success rates (\%) of adversarial attacks against Feature Distillation \cite{DBLP:conf/cvpr/LiuLLXLWW19}, Comdefend \cite{DBLP:conf/cvpr/JiaWCF19}, and Randomized Smoothing \cite{DBLP:conf/icml/CohenRK19} under model ensemble setting. The adversarial examples are generated for the ensemble of Inc-v3, Inc-v4, IncRes-v2, and Res-v2-101 using TI-DIM, NI-TI-DIM, TI-DI-AITM and NI-TI-DI-AITM.}
		\label{t7}
		\begin{tabular}{|c|ccc|c|}
			\hline
			Attack   & Feature Distillation & Comdefend & Randomized Smoothing  & Average  \\
			\hline
			\hline
			TI-DIM     & 83.1 & 78.2 & 49.9 & 70.4\\
			TI-DI-AITM & 90.6 & 87.9 & 63.7 & 80.7\\
			NI-TI-DIM  & 82.1 & 84.7 & 58.6 & 75.1 \\
			NI-TI-DI-AITM  & \textbf{91.4} & \textbf{90.3} & \textbf{66.4} & \textbf{82.7}\\
			\hline
		\end{tabular}
	\end{table*}
	
	\subsection{The comparison of running efficiency}
	\label{s5}
	We compare the running time of each attack mentioned in Table~\ref{table1} using a piece of Nvidia GPU GTX 1080 Ti. Table~\ref{table5} shows the running time under single-model setting and model ensemble setting. It can be seen that attacks combined with our method AI-FGTM do not cost extra running. Additionally, SIM requires at least two pieces of GPUs under model ensemble setting and costs much more running time than other attacks under both single-model setting and model ensemble setting. Therefore, we exclude SIM in the following experiments.
	
	\subsection{Ablation study}
	\label{s444}
	Table~\ref{table_ablation} shows the ablation study of the effects of different parts of our method. We compare the mean perturbation and the mean success rates of the adversarial examples against six classic defense models. Our observations are shown as follow:
	
	\begin{itemize}
		\item[1.] Both of the tanh function and Adam can reduce the perturbation size. Additionally, Adam can also improve the transferability of adversarial examples.
		\item[2.] The combination of the tanh function and Adam can greatly reduce the perturbation size, but only slightly improve the transferability of adversarial examples.
		\item[3.] Using smaller kernels and dynamic step size can improve the transferability of adversarial examples even using dynamic step size slightly increase the perturbation size.
	\end{itemize}
	
	\subsection{The validation results in the single-model attack scenario}
	\label{s4.3}
	
	In this section, we compare the success rates of AI-FGTM based attacks and the baseline attacks against six classic defenses. We generate adversarial attacks for Inc-v3, Incv4, IncRes-v2, and Res-v2-101 by severally using TI-DIM, TI-DI-AITM, NI-TI-DIM and NI-TI-DI-AITM.
	
	As shown in Table~\ref{table3}, we find that our attack method consistently outperforms the baseline attacks by a large margin. Furthermore, according to Table~\ref{table3} and Fig.~\ref{fig:4}(b), we observe that our method can generate adversarial examples with much better transferability and indistinguishability.
	
	\subsection{The validation results in the model ensemble attack scenario}
	\label{s4.4}
	In this section, we present the success rates of adversarial examples generated for an ensemble of four non-defense models. Table~\ref{table4} gives the results of transfer-based attacks against six classic defense models. It shows that our methods achieve higher success rates than baseline attacks. In particular, without extra running time and resource, TI-DI-AITM and NI-TI-DI-AITM fool six defense models with an average success rate of 88.0\% and 89.3\%, respectively, which are higher than the state-of-the-art gradient-based attacks.
	
	We also validate our method by comparing the different results between DIM, TI-DIM and TI-DI-AI-FGTM in Fig.~\ref{fig:4}. Adversarial examples are generated for the ensemble of Inc-v3, Inc-v4, IncRes-v2 and Res-v2-101 using different attack methods. Fig.~\ref{fig:4} (a) shows that the tanh function does not hurt the performance of adversarial examples and Adam can boost the attack success rates. Fig.~\ref{fig:4} (b) shows that our method significantly reduce the mean perturbation size of adversarial examples. In particular, our method reduces 40\% perturbation while delivering the stable performance. Fig.~\ref{fig:4} (c) shows that our approach with $\lambda=1.3$ obtains the largest loss of all the methods.
	
	We evaluate the attacks with three more advanced defenses, namely Feature Distillation \cite{DBLP:conf/cvpr/LiuLLXLWW19}, Comdefend \cite{DBLP:conf/cvpr/JiaWCF19} and Randomized Smoothing \cite{DBLP:conf/icml/CohenRK19}. Table~\ref{t7} shows the success rates of TI-DIM, TI-DI-AITM, NI-TI-DIM and NI-TI-DI-AITM against these defenses in the ensemble attack scenario.
	
	In Table~\ref{t7}, we find that the attacks with AI-FGTM consistently outperform the attacks with MI-FGSM. In general, our methods can fool these defenses with high success rates.
	
	Based on the above experimental results, it is reasonable to state that the proposed TI-DI-AITM and NI-TI-DI-AITM can generate adversarial examples with much better indistinguishability and transferability. Meanwhile, TI-DI-AITM and NI-TI-DI-AITM raise the security challenge for the development of more effective defense models.
	
	\section{Conclusion}
	
	In this paper, we propose AI-FGTM to craft adversarial examples that are much indistinguishable and transferable. AI-FGTM modifies the major gradient processing steps of the basic sign structure to address the limitations faced by the existing basic sign involved methods. Compared with the state-of-the-art attacks, extensive experiments on an ImageNet-compatible dataset show that our method generates more indistinguishable adversarial examples and achieves higher attack success rates without extra running time and resource. Our best attack NI-TI-DI-AITM can fool six classic defense models with an average success rate of 89.3\% and fool three advanced defense models with an average success rate of 82.7\%, which are higher than the state-of-the-art gradient-based attacks. Additionally, our method reduces nearly 20\% mean perturbation. It is expected that our method serves as a new baseline for generating adversarial examples with higher transferability and indistinguishability.
	
\section{Acknowledgements}

	This work was supported in part by National Natural Science Foundation of China under Grant (62106281). This paper was finished with the encouragement of Zou's wife Maojuan Tian. Zou would like thank her and tell her:‘ the most romantic thing I can imagine is gradually getting old with you in scientific exploration.’
		
\bibliography{aaai22}

\section{Appendix}

In this supplementary material, we provide more results in our experiments. In Sec. A, we investigate the effects of different hyper-parameters of AI-FGTM. In Sec. B, we report the success rates in white-box attack to show the effectiveness of AI-FGTM. In Sec. C, we present the visual examples generated by different attacks to show the better indistinguishability of the adversarial examples generated by our methods. In Sec. D, we finally present the value distributions of the accumulated gradients across iterations to demonstrate the limitation of the basic sign attack series. Codes are also available in the supplementary material. Codes are available at https://github.com/278287847/AI-FGTM.

\subsection*{A. The indistinguishability with PSNR and SSIM}

We evaluate the indistinguishability with PSNR and SSIM (higher value indicates better indistinguishability). In detail, We generate adversarial examples for inc-v3 with TI-DIM, TI-DI-AITM, NI-TI-DIM and NI-TI-DI-AITM, and get the mean PSNR and SSIM between the adversarial examples and the clean examples. The results are shown in Table~\ref{t111}. We can find that the attacks with AI-FGTM can always perform better.

\begin{table}[H]
	\centering
	\caption{\label{t111}The indistinguishability evaluated with PSNR and SSIM.}
	\small
	\begin{tabular}{cccccccc}
		\hline
		Attack & PSNR & SSIM\\
		\hline
		\begin{tabular}[c]{@{}c@{}}TI-DIM\\ TI-DI-AITM\end{tabular} & \begin{tabular}[c]{@{}c@{}} 26.81 \\\textbf{28.64}\end{tabular} &\begin{tabular}[c]{@{}c@{}} 0.78 \\\textbf{0.84}\end{tabular} & \\
		\hline
		\begin{tabular}[c]{@{}c@{}}NI-TI-DIM\\ NI-TI-DI-AITM\end{tabular} & \begin{tabular}[c]{@{}c@{}}26.83\\ \textbf{28.63}\end{tabular} &\begin{tabular}[c]{@{}c@{}} 0.77 \\\textbf{0.82}\end{tabular} & \\
		\hline
	\end{tabular}
\end{table}

\subsection*{B. The effects of different hyper-parameters}
\label{s4.2}

\begin{figure*}[!htb]
	\centering
	\includegraphics[width=1\textwidth]{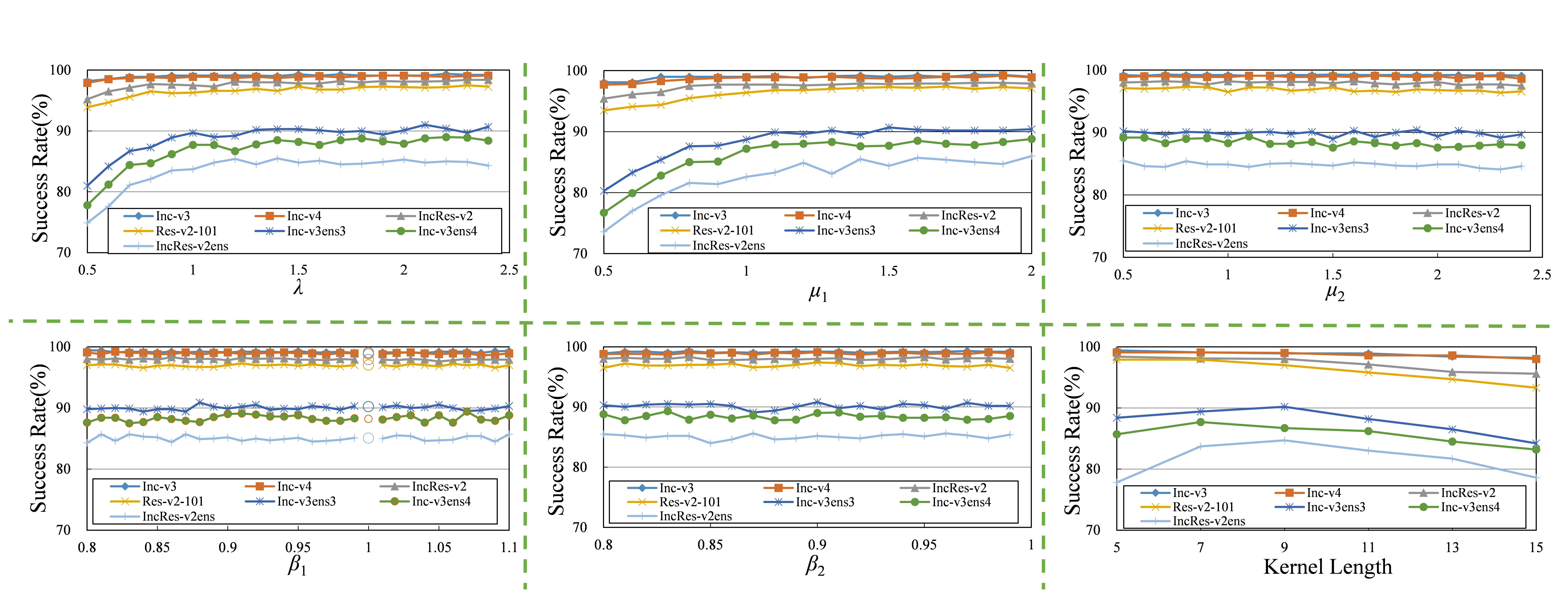}
	\caption{The success rates (\%) of adversarial attacks against Inc-v3, Inc-v4, IncRes-v2, Res-v2-101, Inc-v3ens3, Inc-v3ens4 and IncResv2ens.}
	\label{fig:3}
\end{figure*}

We explore the effects of different hyper-parameters of AI-FGTM and aim to find the appropriate settings to balance the success rates of both white-box and black-box attacks. The adversarial examples are generated for the ensemble of Inc-v3, Inc-v4, IncRes-v2, and Res-v2-101 using TI-DI-AITM. We first show the results of white-box attacks against four known models, and then we present the performance of black-box attacks against three defense models in Fig.~\ref{fig:3}. It can be seen that the appropriate settings are $\lambda = 1.3$, ${\mu _1} = 1.5$, ${\mu _2} = 1.9$, ${\beta _1} = 0.9$, ${\beta _2} = 0.99$, and the kernel length is 9.
\subsection*{C. The effectiveness in the white-box case}

In order to demonstrate the effectiveness of our method, we report the success rates in white-box attack. We present the validation results in single-model attack scenario and model ensemble attack scenario. Table~\ref{t5} presents the success rates of TI-DIM, TI-DI-AITM, NI-TI-DIM, and NI-TI-DI-AITM against white-box models. The details of the abbreviations are stated more precisely in Table 1 of our submitted paper.

As shown in Table~\ref{t5}, we can find that NI-TI-DI-AITM consistently outperform NI-TI-DIM in the single-model attack scenario. Compared with MI-FGSM, experimental results show that our method can improve the performance of NIM. Fig. 2 of our submitted paper also demonstrate that our method can generate adversarial examples with better indistinguishability.

\begin{table*}[b!]
	\centering
	\caption{\label{t5}The success rates (\%) of adversarial attacks against six black-box defense models under single-model setting. The adversarial examples are generated for Inc-v3, Inc-v4, IncRes-v2, Res-v2-101 respectively using NI-TI-DIM and NI-TI-DI-AITM.}
	\small
	\begin{tabular}{cccccccc}
		\hline
		& Attack & Inc-v3ens3 & Inc-v3ens4 & IncRes-v2ens & HGD & R\&P & NIPS-r3\\
		\hline
		Inc-v3 & \begin{tabular}[c]{@{}c@{}}NI-TI-DIM\\ NI-TI-DI-AITM\end{tabular} & \begin{tabular}[c]{@{}c@{}} 48.3 \\\textbf{51.9}\end{tabular} & \begin{tabular}[c]{@{}c@{}}48.6\\ \textbf{52.4}\end{tabular} & \begin{tabular}[c]{@{}c@{}}36.9\\ \textbf{38.2}\end{tabular}
		& \begin{tabular}[c]{@{}c@{}}37.3\\\textbf{39.3}\end{tabular} & \begin{tabular}[c]{@{}c@{}}36.8\\ \textbf{38.1}\end{tabular} & \begin{tabular}[c]{@{}c@{}}42.5\\\textbf{44.6}\end{tabular}\\
		\hline
		Inc-v4 & \begin{tabular}[c]{@{}c@{}}NI-TI-DIM\\ NI-TI-DI-AITM\end{tabular} & \begin{tabular}[c]{@{}c@{}}52.4\\ \textbf{54.8}\end{tabular} & \begin{tabular}[c]{@{}c@{}}51.8\\ \textbf{53.7}\end{tabular} & \begin{tabular}[c]{@{}c@{}} 41.3\\  \textbf{41.7}\end{tabular}
		& \begin{tabular}[c]{@{}c@{}}  41.9\\  \textbf{43.9}\end{tabular} & \begin{tabular}[c]{@{}c@{}}  41.1\\  \textbf{43.2}\end{tabular} & \begin{tabular}[c]{@{}c@{}}  42.7\\  \textbf{44.1}\end{tabular}\\
		\hline
		IncRes-v2 & \begin{tabular}[c]{@{}c@{}}  NI-TI-DIM\\  NI-TI-DI-AITM\end{tabular} & \begin{tabular}[c]{@{}c@{}}  61.5\\  \textbf{66.5}\end{tabular} & \begin{tabular}[c]{@{}c@{}}60.4 \\ \textbf{63.8}\end{tabular} & \begin{tabular}[c]{@{}c@{}} 59.9\\  \textbf{62.0}\end{tabular}
		& \begin{tabular}[c]{@{}c@{}} 60.1\\  \textbf{63.2}\end{tabular} & \begin{tabular}[c]{@{}c@{}} 62.2 \\\textbf{65.6}\end{tabular} & \begin{tabular}[c]{@{}c@{}} 63.1\\  \textbf{65.8}\end{tabular}\\
		\hline
		Res-v2-101 & \begin{tabular}[c]{@{}c@{}}  NI-TI-DIM\\  NI-TI-DI-AITM\end{tabular} & \begin{tabular}[c]{@{}c@{}} 59.5\\  \textbf{64.0}\end{tabular} & \begin{tabular}[c]{@{}c@{}} 57.7\\  \textbf{61.0}\end{tabular} & \begin{tabular}[c]{@{}c@{}} 50.4\\  \textbf{54.6}\end{tabular}
		& \begin{tabular}[c]{@{}c@{}}  51.9\\  \textbf{54.8}\end{tabular} & \begin{tabular}[c]{@{}c@{}}  50.8\\  \textbf{53.4}\end{tabular} & \begin{tabular}[c]{@{}c@{}}  54.6\\  \textbf{57.6}\end{tabular}\\
		\hline
	\end{tabular}
\end{table*}



\subsection*{D. Visualization of adversarial examples}

We visualize six groups of adversarial examples generated by six different attacks in Fig~\ref{fig:4}. The adversarial examples are crafted on the ensemble models, including Inc-v3, Inc-v4, IncRes-v2 and Res-101, using TI-DIM, TI-DI-AITM ($\lambda$=1.3), TI-DI-AITM ($\lambda$=0.65), NI-TI-DIM, NI-TI-DI-AITM ($\lambda$=1.3) and NI-TI-DI-AITM ($\lambda$=0.65). We see that adversarial examples generates by our method have the better indistinguishability.

\begin{figure*}[t]
	\centering
	\includegraphics[width=1\textwidth]{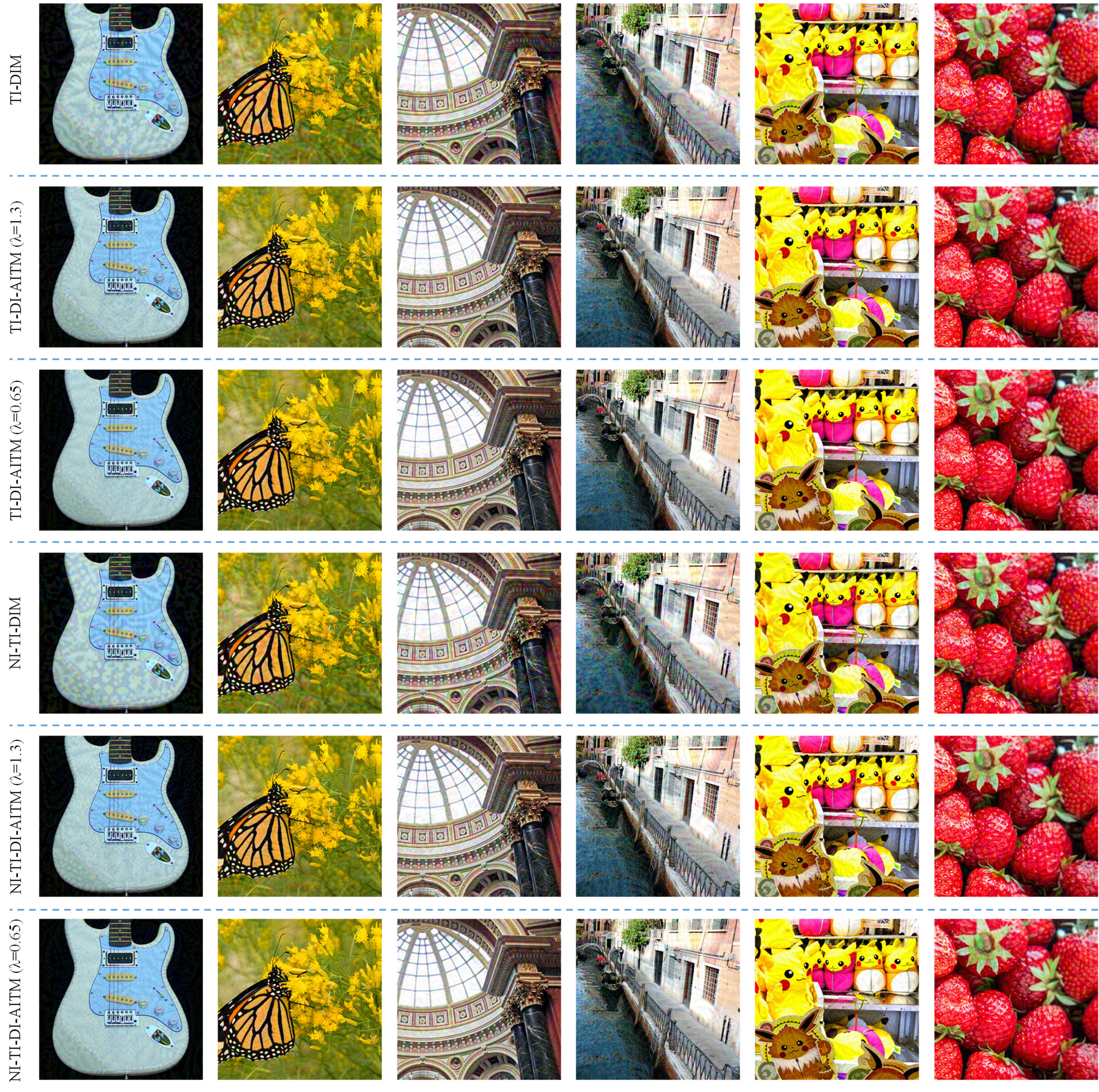}
	\caption{Visualization of six groups of adversarial examples generated by six different attacks.}
	\label{fig:4}
\end{figure*}

\subsection*{E. The value distributions of the accumulated gradients across iterations}

\begin{figure*}[t]
	\centering
	\includegraphics[width=1\textwidth]{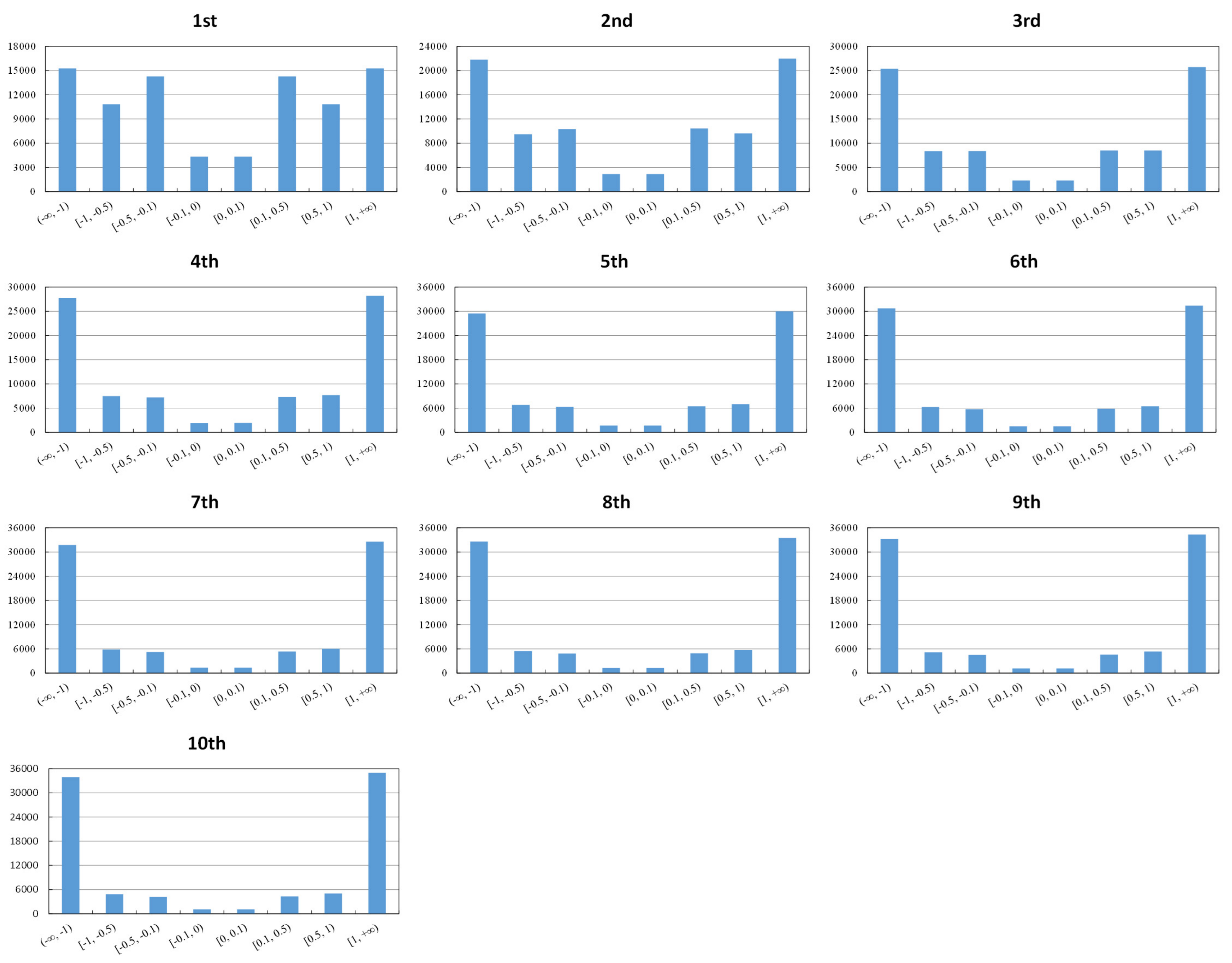}
	\caption{The value distributions of the accumulated gradients across iterations.}
	\label{fig:15}
\end{figure*}

We present the value distributions of the accumulated gradients across iterations to demonstrate the limitation of the basic sign attack series. With iteration number $T=10$, we present the value distributions of the accumulated gradients in each iteration in Fig~\ref{fig:15}. As the number of iterations increases, values of the accumulated gradients tend to be greater than 1 or less than -1. However, a large number of values are in the range of $(-0.5, 0.5)$ in the first three iterations. With constant step size ${\rm{\alpha }} = \varepsilon /T$, perturbation size of adversarial examples is greatly enlarged. Hence, we replace the sign function with the tanh function and gradually increase the step size.

\end{document}